\documentclass[letterpaper, 10 pt, conference]{ieeeconf}  

\IEEEoverridecommandlockouts                              

\AtEndDocument{\par\leavevmode} 

\usepackage[draft]{pgf}
\usepackage{cite}
\usepackage{flushend}
\usepackage{balance}

\usepackage{graphics} 
\usepackage{graphicx} 
\usepackage{epsfig} 
\usepackage{amsmath, bm} 
\usepackage{amssymb}  
\usepackage{upgreek}

\usepackage{cuted}
\usepackage{booktabs}
\usepackage{multicol}
\usepackage{multirow}
\usepackage{stfloats}
\usepackage{lipsum}  
\usepackage{breqn}
\usepackage{comment} 
\usepackage{accents}

\newcommand\thickbar[1]{\accentset{\rule{.4em}{.8pt}}{#1}}

\title{\LARGE \bf
Enforcing nonholonomic constraints in \textit{Aerobat},\\
a roosting flapping wing model
}

\author{Eric Sihite, and Alireza Ramezani, {\it SiliconSynapse Laboratory}$^{1}$
\thanks{$^{1}$ECE Department, Northeastern University, Boston, MA, USA.}
\thanks{emails: \{e.sihite, a.ramezani\}@northeastern.edu}%
}

\begin{document}

\maketitle

\global\csname @topnum\endcsname 0
\global\csname @botnum\endcsname 0

\thispagestyle{empty}
\pagestyle{empty}

\begin{abstract}


Flapping wing flight is a challenging dynamical problem and is also a very fascinating subject to study in the field of biomimetic robotics. A Bat, in particular, has a very articulated armwing mechanism with high degrees-of-freedom and flexibility which allows the animal to perform highly dynamic and complex maneuvers, such as upside-down perching. This paper presents the derivation of a multi-body dynamical system of a bio-inspired bat robot called \textit{Aerobat} which captures multiple biologically meaningful degrees-of-freedom for flapping flight that is present in biological bats. Then, the work attempts to manifest closed-loop aerial body reorientation and preparation for landing through the manipulation of inertial dynamics and aerodynamics by enforcing nonholonomic constraints onto the system. The proposed design paradigm assumes for rapidly exponentially stable controllers that enforce holonomic constraints in the joint space of the model. A model and optimization-based nonlinear controller is applied to resolve the joint trajectories such that the desired angular momentum about the roll axis is achieved.

\end{abstract}


\section{Introduction}
\label{sec:introduction}

\begin{figure}
    \vspace{0.05in}
    \centering
    \includegraphics[width=\linewidth]{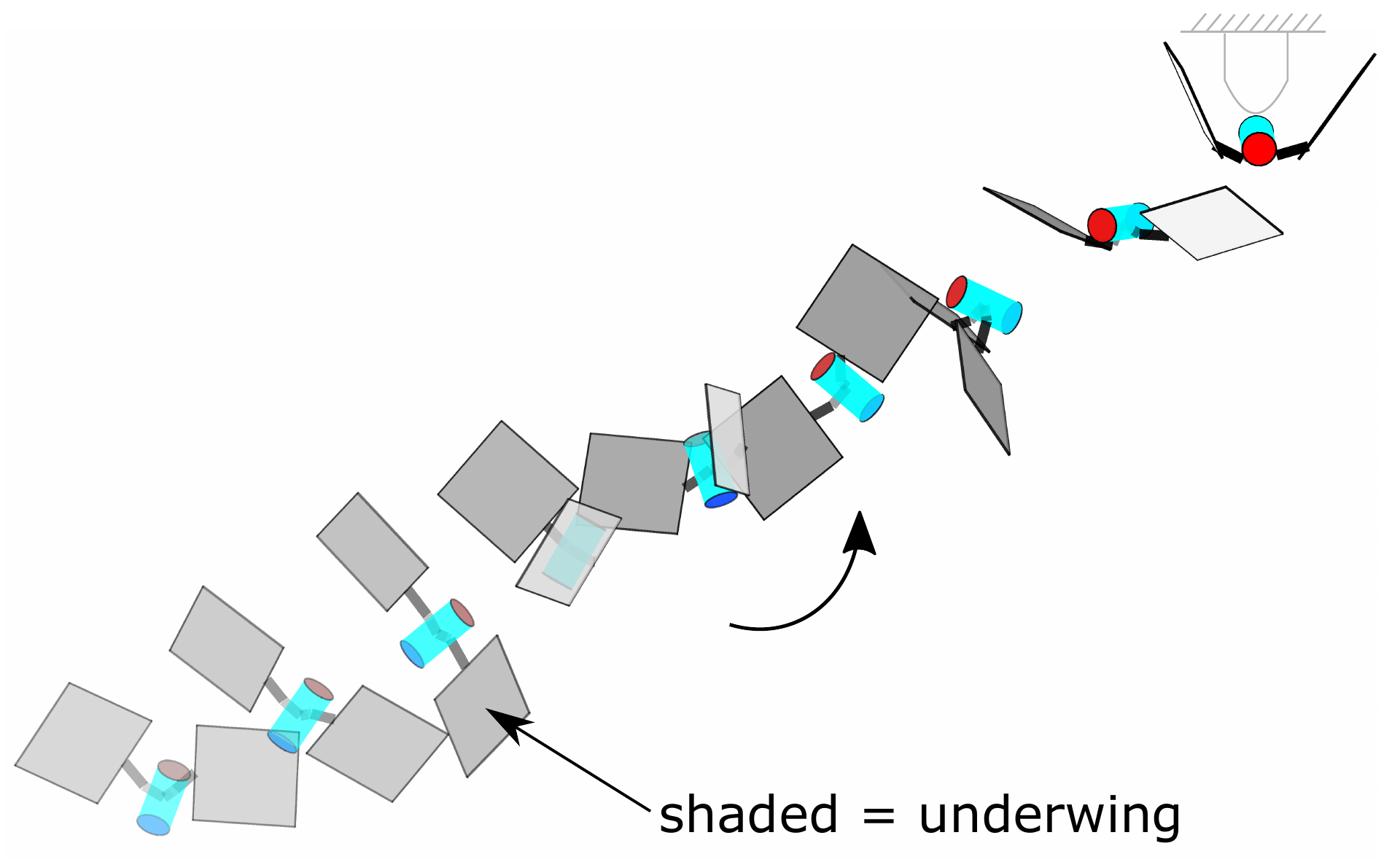}
    \caption{The simulated upside-down perching maneuver of \textit{Aerobat}.}
    \label{fig:aerobat_flip}
    \vspace{-0.05in}
\end{figure}

The key hypothesis this work tries to inspect is: ''The feasibility of performing flip turns, a well known attribute of a bat-like landing maneuver, through the manipulation of inertial dynamics while including aerodynamic forces.'' Bats (and birds) possess no energy-hungry motors widely used in flying robots for thrust vectoring yet they are more capable than any of these systems when agility and energy efficiency of flight are concerned. Flying vertebrates apply the combination of inertial dynamics and aerodynamics manipulations to showcase extremely agile maneuvers. Unlike rotary- and fixed-wing systems wherein aerodynamic surface (i.e., ailerons, rudders, propellers, etc.) come with the sole purpose of aerodynamic force adjustment, the wings (also called appendages) in birds and bats possess more sophisticated roles. It is known that birds perform zero-angular-momentum turns by making differential adjustments (e.g., collapsing armwings) in the inertial forces led by one wing versus the other. Bats apply a similar mechanism to perform sharp banking turns \cite{riskin2012upstroke}.

Among these maneuvers, landing (or perching), which flying vertebrates do it in one way or another for a variety of reasons (e.g., transition to walking, resting on a perch, hanging from the ceiling of a cave, etc.), is an interesting maneuver to take inspiration from for aerial robot designs. Perching birds rotate their wings so that the aerodynamic drag is increased by creating a high-pressure region inside of the wings and a low-pressure region behind the wings. This brings the wings to a stalled condition at which point the generated lift is equal to zero and the animal falls naturally while employing the legs as a landing gear. Bats do it in a radically different way. After the self-created stalled condition, they manifest an acrobatic heels-above-head maneuver that involves catapulting the lower body in a similar way that a free style swimmer flip turns. Perching insects and birds have been the source of inspiration and bio-mimicry of them has led to interesting robot designs in recent years \cite{paranjape2013novel,doyle2012avian,graule2016perching}. Remarkably, the bio-mimicry of bat-like landing is overlooked mainly because not only the aerodynamics adjustments are involved but also unique design provisions are required to allow for the manipulation of inertial dynamics. 

A bat-style landing maneuver is extremely rich in dynamics and control and its characteristics are overlooked. Much of attention has been paid to simpler dynamics such as hovering and straight flight. While mathematical models of insect-style, rotary- and fixed-wing robots of varying size and complexity are relatively well developed, models of airborne, fluidic-based vertebrates locomotion remain largely open due to the complex body articulation involved in their flight. The mainstream school of thought inspired by insect flight has conceptualized wing as a mass-less, rigid structure, which is nearly planar and translates -- as a whole or in two-three rigid parts -- through space \cite{bergou2010fruit, hedrick2009wingbeat}. In this view, wings possess no inertial effect, are fast that yield two-time-scale dynamics, permit quasi-static external force descriptions, and tractable dynamical system. Unfortunately, these paradigms fail to provide insight into airborne, vertebrate locomotion and an ingredient of a more complete and biologically realistic model is missing, that is, the manipulation of inertial dynamics. The manipulation of inertial dynamics is an under-appreciated aspect in existing paradigms.

The objective of this work is to manifest closed-loop aerial body reorientation and preparation for landing through the manipulation of inertial dynamics and aerodynamics by enforcing nonholonomic constraints onto the system. The proposed design paradigm assumes for rapidly exponentially stable controllers that enforce holonomic constraints in the joint space of the model. Then, a model and optimization-based nonlinear controller is applied to resolve the joint trajectories such that the desired angular momentum about the roll axis is achieved. First, a brief overview of our motivation will be presented followed by a model description of the landing maneuver. Then, a dynamic model and optimization problem will be derived in detail and the preliminary simulation results will be reported at the end followed by final remarks and conclusion.



\begin{figure*}[t]
    \vspace{0.05in}
    \centering
    \includegraphics[width=0.9\linewidth]{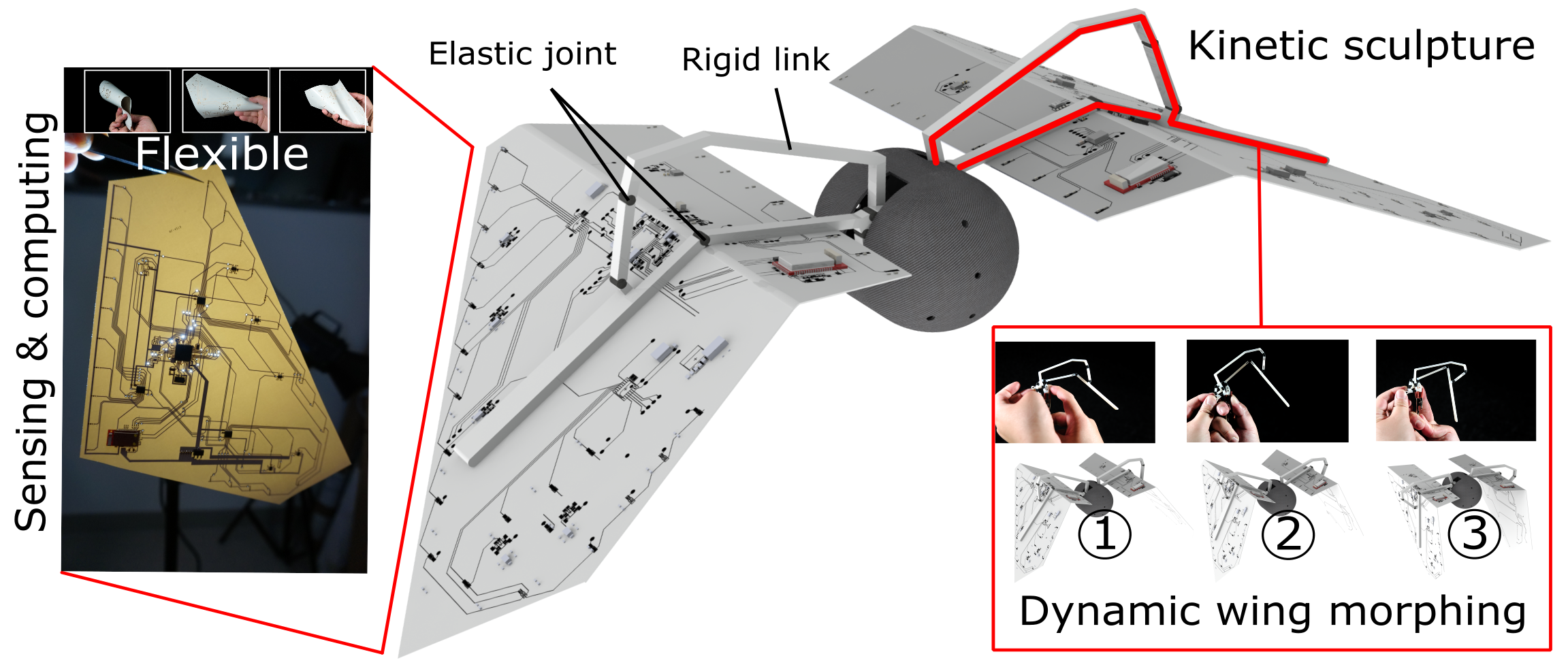}
    \caption{The bio-inspired bat robot, called \textit{Aerobat}, uses monolithically fabricated rigid and soft arm-armwing structure. The white material of the armwing is rigid while the black material is a flexible living hinge. The flexible PCB forms the wing membrane which helps reducing the overall weight of the robot and the \textit{kinetic sculpture} forms a compliant linkage mechanism that can conform and articulate flapping motion.}
    \label{fig:satan_overview}
    \vspace{-0.05in}
\end{figure*}

Fig. \ref{fig:satan_overview} shows the concept design of our bio-inspired bat robot, called \textit{Aerobat} \cite{sihite2020computational}, which is a continuation to our previous work in \cite{ramezani2016bat, hoff2016synergistic, hoff2017reducing, ramezani_describing_2017, ramezani2017biomimetic, syed_rousettus_2017, hoff2018optimizing}. The control and dynamic modeling of this robot were also investigated in \cite{ramezani_lagrangian_2015,ramezani_modeling_2016,ramezani_nonlinear_2016,hoff_trajectory_2019}. This bioinspired robot is designed to be very lightweight and the flexible armwing mechanism is designed to mimic some of the biologically meaningful degrees-of-freedom (DoF) in a bat's flapping gait. This flexible bat armwing structure, called the \textit{kinetic sculpture}, uses both rigid and flexible materials that is monolithically fabricated using PolyJet 3D printing technology. This armwing is articulated through a series of four-bar linkages and crank mechanism which is driven by a single motor to articulate the wing expansion and retraction during downstroke and upstroke respectively. This results in maximum wingspan and lift during the downstroke while reducing the negative lift during the upstroke, therefore forming an efficient flapping gait. We have also developed a launching landing apparatus called \textit{Harpoon} as shown in \cite{ramezani2020towards} which will be used in the actual robot once it has successfully performed the upside-down maneuver to latch onto the perching location. 


The motivation of developing Aerobat is to mimic the complex and highly articulated natural bat armwing by deforming the armwing morphology and achieve a varying range of motion that mimics the flapping motion of a biological bat. This means of articulation facilitates control through \textit{morphological computation} \cite{hauser2011towards}, where a simple control action can actuate a very complex motion or trajectory which is suitable for mimicking the complex wing articulation of an actual bat. This motivates us to develop a dynamic model for simulations which can be used to develop the stabilizing controller for this bat robot. As part of the future work of the current armwing design, we will extend the biologically meaningful degrees-of-freedom of the armwing to include the feathering and mediolateral movements. This work aims to find the optimal gait for this armwing structure to achieve stable flight and other complex maneuvers, such as the upside-down perching maneuver.

\section{System Dynamic Modeling}
\label{sec:modeling}

\begin{figure}
    \vspace{0.05in}
    \centering
    \includegraphics[width=\linewidth]{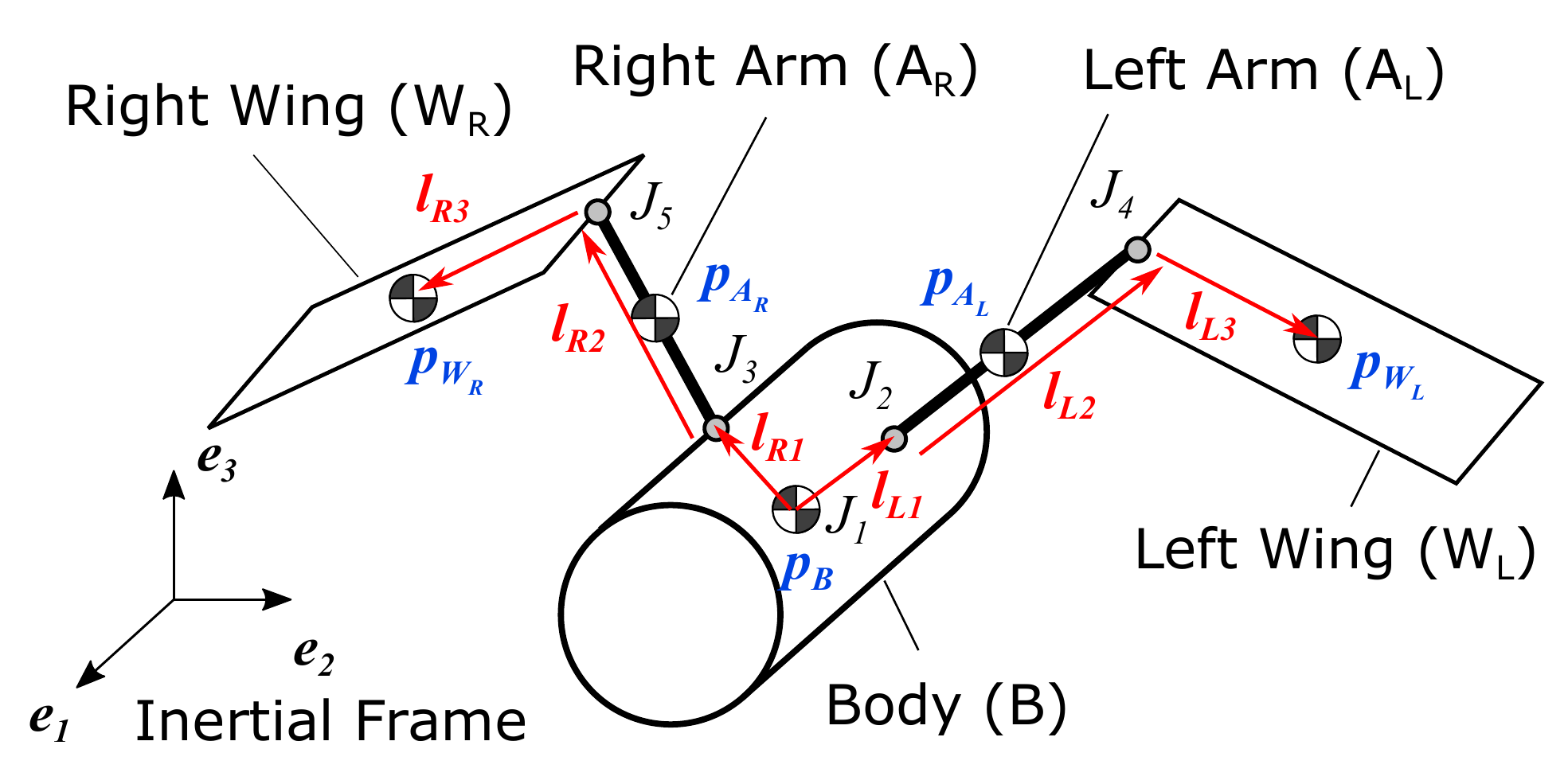}
    \caption{Aerobat model using five rotating bodies where joints $J_i, i=\{1,\dots,5\}$ represent the center of rotations, the linear position $\bm{p}$ represents the center of mass, and $\bm{l}$ represents the length vectors relevant to the Aerobat model conformation.}
    \label{fig:aerobat_model}
    \vspace{-0.05in}
\end{figure}

\begin{figure}
    \vspace{0.05in}
    \centering
    \includegraphics[width=0.7\linewidth]{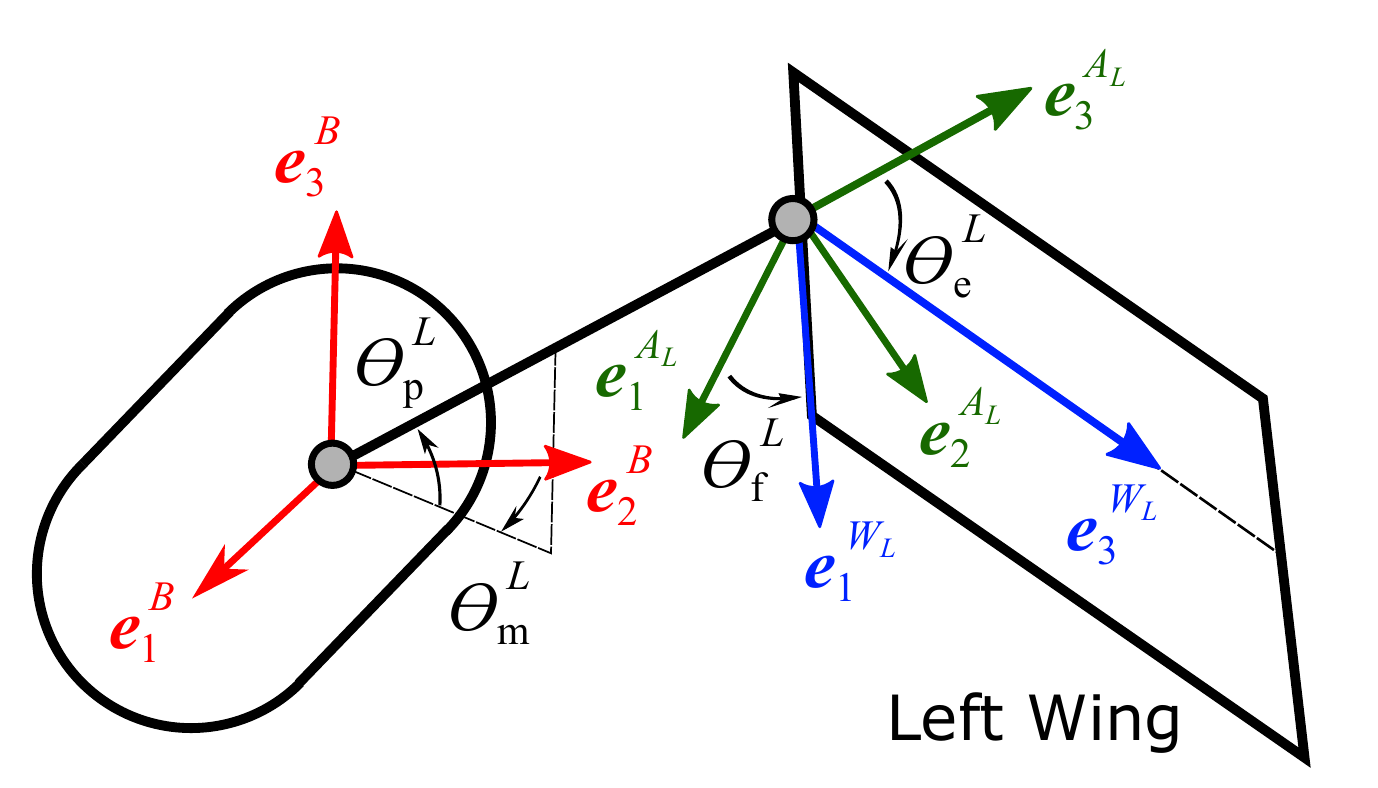}
    \caption{Aerobat biologically meaningful degrees-of-freedom angles: armwing plunging ($\theta_p$), mediolateral ($\theta_m$), elbow extension ($\theta_e$), and feathering ($\theta_f$). $\theta_p$ and $\theta_m$ are defined with respect to the body frame while $\theta_e$ and $\theta_f$ are defined with respect to the arm frame. The superscript $L$ and $R$ are used to represent left and right wing angles respectively.}
    \label{fig:aerobat_dof}
    \vspace{-0.05in}
\end{figure}

The Aerobat system can be modeled as five rotating bodies which are attached to one another with joints or hinges, as shown in Fig. \ref{fig:aerobat_model}. This armwing mechanism is designed to follow the biologically meaningful degrees-of-freedom (DoF) of a bat's armwing, as illustrated in Fig. \ref{fig:aerobat_dof}, where these DoFs are important for flapping wing flight. The shoulder joint plunge angle $(\theta_p)$ controls the wing upstroke and downstroke motion which forms the core flapping motion. The elbow extension/flexion angle $(\theta_e)$ expands the wing during the downstroke and retracts it during the upstroke motion, which improves efficiency by reducing the negative lift during the upstroke. The mediolateral motion $(\theta_m)$ extends the wings forward and the feathering motion $(\theta_f)$ rotates the wing surface plane with respect to the arm which has an effect of changing the angle of attack.

\subsection{Euler-Lagrangian Dynamic Formulation}
\label{ssec:modeling_dynamic}

Let a vector with a superscript notation represents the vector defined in a non-inertial coordinate frame and the vector without superscript is defined about the inertial frame, e.g. $\bm{x}^{B}$ is the vector $\bm{x}$ about frame $B$. The frames of references is illustrated in Fig. \ref{fig:aerobat_model} and the coordinate frame rotation of the five bodies can be defined as follows:
\begin{equation}
    \begin{aligned}
        \bm{x} &= R_B\,\bm{x}^{B}, \quad &
        \bm{x}^B &= R_{A_L}\,\bm{x}^{A_L} = R_{A_R}\,\bm{x}^{A_R} \\
        \bm{x}^{A_L} &= R_{W_L}\,\bm{x}^{W_L}, \quad &
        \bm{x}^{A_R} &= R_{W_R}\,\bm{x}^{W_R},
    \end{aligned}
\label{eq:rotation_matrices}
\end{equation}
where $R_B$ is the body rotation matrix about the inertial frame, $R_{A_L}$ and $R_{A_R}$ are the rotation matrix of the left and right arm respectively about the body frame, and $R_{W_L}$ and $R_{W_R}$ are the rotation matrix of the left and right wing respectively about their respective arm. The corresponding angular velocities for these rotation matrices must also be represented in the appropriate coordinate frames. Then we have the following rotation matrix and angular velocity pairs: $(R_B, \bm{\omega}_B)$, $(R_{A_L}, \bm{\omega}_{A_L}^{B})$, $(R_{A_R}, \bm{\omega}_{A_R}^{B})$, $(R_{W_L}, \bm{\omega}_{W_L}^{A_L})$, and $(R_{W_R}, \bm{\omega}_{W_R}^{A_R})$. Let $\theta$ angles be the biologically meaningful flapping angles, as illustrated in Fig. \ref{fig:aerobat_dof}, and the superscript $L$ and $R$ represents the left and right wing joint angles respectively. Then the left armwing rotation matrices are defined as follows:
\begin{equation}
    \begin{aligned}
        R_{A_L} &= R_z(\theta^L_m)\,R_x(\theta^L_p),\, &
        R_{W_L} &= R_x(\theta^L_e)\,R_z(\theta^L_f),
    \end{aligned}
\label{eq:rotation_matrices_left}
\end{equation}
where $R_x(\theta)$ and $R_z(\theta)$ are the rotational matrix about $x$ and $z$ axis respectively. The left armwing angular velocities are defined as follows:
\begin{equation}
    \begin{aligned}
        \bm{\omega}_{A_L}^{B} &= \begin{bmatrix} 0, 0, \dot{\theta}_m^L \end{bmatrix}^\top + R_z(\theta^L_m)  \begin{bmatrix} \dot{\theta}_p^L, 0 , 0 \end{bmatrix}^\top + \bm{\omega}_{B}^{B}\\
        \bm{\omega}_{W_L}^{A_L} &= \begin{bmatrix} 
        \dot{\theta}_e^L, 0 , 0 \end{bmatrix}^\top + R_x(\theta^L_e) 
        \begin{bmatrix} 0, 0, \dot{\theta}_f^L \end{bmatrix}^\top + \bm{\omega}_{A_L}^{A_L}.
    \end{aligned}
\label{eq:angular_vel_left}
\end{equation}
The right wing derivations can be derived in a similar fashion. Therefore, for the rest of the paper, only the left wing components will be derived if the right side also follow a similar derivation.

As shown in Fig. \ref{fig:aerobat_model}, let $\bm p_B$ be the linear position of the center of mass of a body, $\bm{l}_{Lj}$ and $\bm{l}_{Rj}$, $j = \{1,2,3\}$, be the length vectors which represent the Aerobat mechanism morphology that are constant with respect to their local frame of reference. Then the linear position of the center of mass of the left armwing can be derived as follows:
\begin{equation}
    \begin{aligned}
        \bm{p}_{A_L} &= \bm{p}_B + R_{B}\,\bm{l}_{L1}^{B} + \tfrac{1}{2}\,R_B\,R_{A_L}\,\bm{l}_{L2}^{A_L} \\
        \bm{p}_{W_L} &= \bm{p}_{A_L} + \tfrac{1}{2}\,R_B\,R_{A_L}\,\bm{l}_{L2}^{A_L} + R_B\,R_{A_L}\,R_{W_L}\,\bm{l}_{L3}^{W_L},
    \end{aligned}
\label{eq:linear_position}
\end{equation}
The linear velocity of the center of mass can be derived from \eqref{eq:linear_position} by differentiating the linear positions with respect to time. Note that the linear positions and velocities are defined with respect to the inertial frame. 


The kinetic and potential energy of the system can be derived as follows:
\begin{equation}
    \begin{aligned}
    T &= \sum_{F \in \mathcal{F}} \left( m_F\,\dot{\bm{p}}_F^\top\,\dot{\bm{p}}_F + (\bm{\omega}_F^{F})^\top\,\hat{I}_{F}\,\bm{\omega}_F^{F} \right)\frac{1}{2} \\
    U &= \sum_{F \in \mathcal{F}} m_F \, [ 0, 0, g ] \, \bm{p}_F,
    \end{aligned}
\label{eq:energy}
\end{equation}
where $\mathcal{F} = \{B, A_L, A_R. W_L, W_R\}$ is the set containing the frame of references, $m_F$ and $\hat{I}_{F}$ are the mass and inertia matrix of the corresponding body respectively. $\hat{I}_{F}$ is defined about the local frame of reference which is diagonal and constant. Then the Lagrangian of the system, $L = T - U$, can be used to derive the equation of motion. 

The body rotation $(R_B,\omega_B)$ is derived using the modified Euler-Lagrangian formulation for a rotation in SO(3). This formulation is not susceptible to gimbal lock which might happen if we use Tait-Bryan angles during the upside-down maneuver. The modified Euler-Lagrange equation for rotation in SO(3) can be derived by using Hamilton's principle \cite{lee2017global}, which has the following form:
\begin{equation}
\begin{gathered}
    \frac{d}{dt}\frac{\partial L}{\partial \bm{\omega}_B^B} + \bm{\omega}_B^B \times \frac{\partial L}{\partial \bm{\omega}_B^B} + \sum^3_{j=1}\bm{r}_{B,j} \times \frac{\partial L}{\partial \bm{r}_{Bj}} = \bm{\tau}_B^B \\
    \dot{R}_B = R_B\,S(\bm{\omega}_B^B),
\end{gathered}
\label{eq:eom_hamiltonian}
\end{equation}
where $S(\cdot)$ is a skew operator, $R_B^\top = [\bm{r}_{B1}, \bm{r}_{B2}, \bm{r}_{B3}]$ and $\bm{\tau}_B^B$ is the non-conservative torque about the generalized coordinate $\bm{\omega}_B^B$. The equation of motion of the remaining states can be solved by using the Euler-Lagrange equation:
\begin{equation}
    \begin{gathered}
        \bm{\theta}_L = [\theta_p^L, \theta_m^L, \theta_e^L, \theta_f^L], \qquad \bm{\theta}_R = [\theta_p^R, \theta_m^R, \theta_e^R, \theta_f^R], \\
        \bm{q}_{e} = [\bm{p}_B^\top, \bm{\theta}_L^\top, \bm{\theta}_R^\top]^\top, \qquad
        \frac{d}{dt}\frac{\partial L}{\partial \dot{\bm{q}}_{e}} - \frac{\partial L}{\partial \bm{q}_{e}} = \bm{u}_{e},
    \end{gathered}
\label{eq:eom_euler_lagrangian}
\end{equation}
where $\bm{u}_{e}$ is the non-conservative force about the generalized coordinate $\bm{q}_{e}$. Combining \eqref{eq:eom_hamiltonian} and \eqref{eq:eom_euler_lagrangian}, the equation of motion can be formulated into the following form:
\begin{equation}
    \begin{gathered}
    \bm{q} = [\bm{r}_B^\top, \bm{p}_B^\top, \bm{\theta}_L^\top, \bm{\theta}_R^\top]^\top, \qquad
    \bm{q}_d = [ \bm{\omega}_{B}^\top, \dot{\bm{p}}_B^\top, \dot{\bm{\theta}}_L^\top, \dot{\bm{\theta}}_R^\top]^\top \\
    M\,\dot{\bm{q}}_d + \bm{h} = B_a\, \bm{u}_a + B_m\, \bm{u}_m,
    \end{gathered}
\label{eq:eom_summary}
\end{equation}
where $\bm{r}_B$ is the rotation matrix $R_B$ concatenated into a vector form, $\bm{u}_a$ is the generalized aerodynamic forces and torque. $\bm{u}_m$ is the generalized motor torque acting on the armwing joints which is selected to directly actuate the joints angles $\bm \theta_L$ and $\bm \theta_R$:
\begin{equation}
    \begin{aligned}
    \bm{u}_m &= [\tau_p^L, \tau_m^L, \tau_e^L, \tau_f^L, \tau_p^R, \tau_m^R, \tau_e^R, \tau_f^R]^\top \\
    B_m &= [0_{8 \times 6}, I_{8 \times 8}]^\top,
    \end{aligned}
\end{equation}
where $\tau$ represents the torque acting on each joints.


\subsection{Aerodynamic Modeling}
\label{ssec:modeling_aerodynamic}

\begin{figure}[t]
    \vspace{0.05in}
    \centering
    \includegraphics[width=0.8\linewidth]{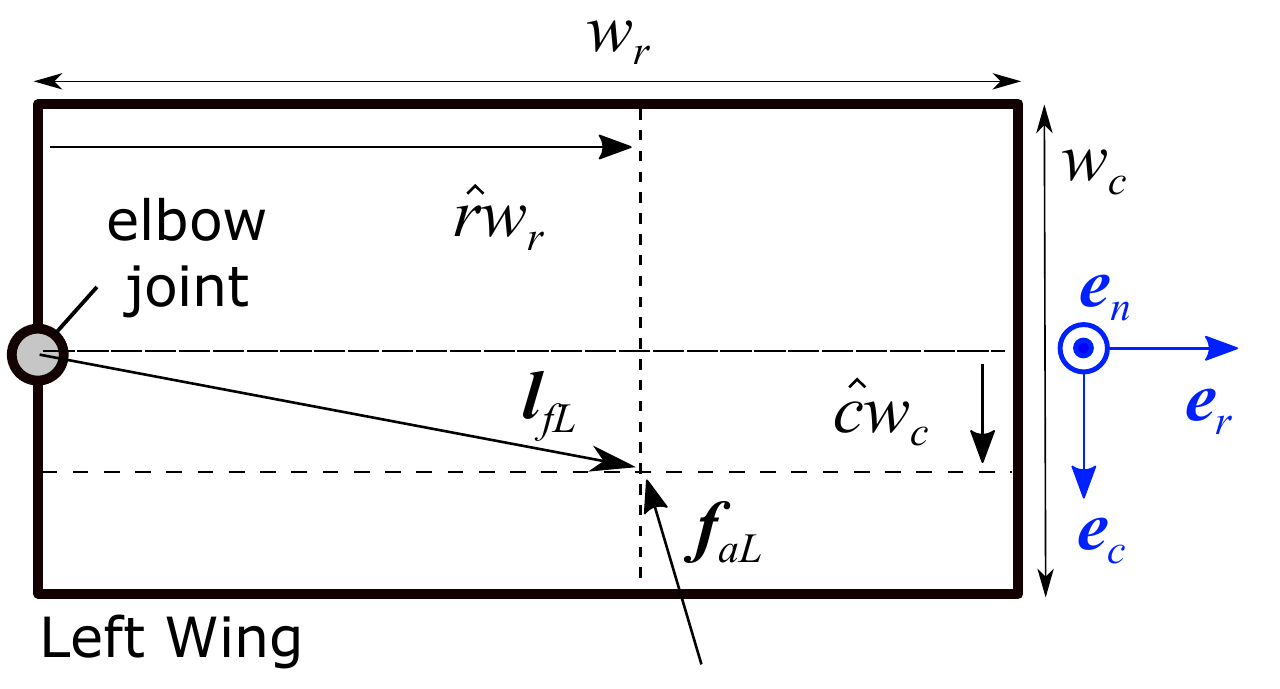}
    \caption{The coordinate system of the wing plate. $w_r$ and $w_c$ are the wing span and chord length respectively. $\hat{r} \in [0,1]$ and $\hat{c} \in [-0.5,0.5]$ are the unitless variables to represent a position on the wing surface. $\bm{l}_{f_L}$ is the length vector to the aerodynamic force $\bm{f}_{a_L}$ acting on the left wing.}
    \label{fig:wing_aerodynamics}
    \vspace{-0.05in}
\end{figure}

The aerodynamic forces can be derived using the virtual displacement defined at the position of the applied aerodynamics force $\bm f_{a_L}$. Let $\bm{p}_{f_L}$ be the position of the applied aerodynamics force $\bm u_L$ on the left wing, defined as follows:
\begin{equation}
    \begin{aligned}
        \bm{p}_{f_L} &= \bm{p}_{A_L} + \tfrac{1}{2}\,R_B\,R_{A_L}\,\bm{l}_{L2}^{A_L} + R_B\,R_{A_L}\,R_{W_L}\,\bm{l}_{f_L}^{W_L}
    \end{aligned}
\label{eq:aero_virtualdisp}
\end{equation}
where $\bm{l}_{f_L}^{W_L}$ is the length vector from the left wing elbow joint to where $\bm{f}_{a_L}$ is applied, as illustrated in Fig. \ref{fig:wing_aerodynamics}. Define the position on the wing surface as follows:
\begin{equation}
    \bm{l}_{f_L} = \hat{r}\,w_r \, \bm{e}_r + \hat c \,w_c \, \bm{e}_c
\end{equation}
where $w_r$ and $w_c$ are the wing span and chord length respectively. The position is represented using the unitless variables $\hat r \in [0, 1]$ and $\hat c \in [-0.5, 0.5]$, as illustrated in Fig. \ref{fig:wing_aerodynamics}. For the left wing, $\{ \bm e_c, \bm e_n, \bm e_r  \} = \{ \bm e_1^{W_L},  -\bm e_2^{W_L}, \bm e_3^{W_L}  \}$. Then $\bm p_{f_L}$ and $\bm p_{f_R}$ can be represented using the coordinate $(\hat c, \hat r)$, e.g. $\bm p_{f_L} (\hat c, \hat r)$.


The generalized forces can be derived by as follows:
\begin{equation}
    Q_j = \bm{f}_{a_L}^\top\,\frac{\partial \dot{\bm{p}}_{f_L}}{\partial q_{d,j}} = \left( \frac{\partial \dot{\bm{p}}_{f_L}}{\partial q_{d,j}} \right)^\top R_B\,R_{A_L}\,R_{W_L}\,\bm{f}_{a_L}^{W_L},
    \label{eq:generalized_forces}
\end{equation}
where $q_{d,j}$ is the $j$'th component of the vector $\bm{q}_d$ defined in \eqref{eq:eom_summary}. Since this is a flapping wing robot, the airfoil speed is variable across the wingspan which means that we need to integrate the aerodynamic forces across the wing surface. It is possible to represent the $B_a$ matrix without $\bm{l}_{f_L}^{W_L}$ by calculating the aerodynamic torque $\bm{\tau}_{a_L}^{A_L}$ about the plate joint. Evaluate the component of  \eqref{eq:generalized_forces} that has the $\bm{l}_{f_L}^{W_L}$ term using the following method:
\begin{equation}
\begin{aligned}
    Q_{\tau_j} & = 
    \bm{f}^\top_{a_L}\frac{\partial}{\partial q_{d,j}}\left( R_B\,R_{A_L}\, (\bm{\omega}_{W_L}^{A_L} \times R_{W_L}\,\bm{l}_{f_L}^{W_L}) \right) \\
    & = \frac{\partial \bm{\omega}_{W_L}^{A_L}}{\partial q_{d,j}}^\top  \mathrm{cof}(R_{W_L})(\bm{l}_{f_L}^{W_L} \times \bm{f}_{a_L}^{W_L}),
\end{aligned}
\label{eq:generalized_torque}
\end{equation}
where $\bm{\tau}_{a_L}^{W_L} = \bm{l}_{f_L}^{W_L} \times \bm{f}_{a_L}^{W_L}$ is the aerodynamic torque acting on the left wing plate about its joint. The right wing also follows a similar derivation, then combining \eqref{eq:generalized_forces} and \eqref{eq:generalized_torque} for both sides of the wing forms the matrix $B_a \in \mathcal{R}^{14 \times 12}$ with the input vector $\bm{u}_a = [\bm{f}_{a_L}^{W_L \top}, \bm{\tau}_{a_L}^{W_L \top}, \bm{f}_{a_R}^{W_R \top}, \bm{\tau}_{a_R}^{W_R \top}]^\top$ for the aerodynamic forces and torques acting on the wings which must be integrated about the wing surface. The integration can be solved using the following equations:
\begin{equation}
\begin{aligned}
    \bm{l}_{f_L} (\hat{r}) &= \hat{r}\,w_r \, \bm{e}_r + 0.25\,w_c \, \bm{e}_c \\
    \alpha (\hat{r}) &= \mathrm{atan2}(\bm{e}_n^\top\,\bm{v}_w (\hat{r}), \bm{e}_c^\top\,\bm{v}_w (\hat{r})) \\
    f_l(\hat{r}) &=  \rho\,w_c w_r |\bm{v}_w|^2 C_L(\alpha (\hat{r}))/2 \\
    f_d(\hat{r}) &=  \rho\,w_c w_r |\bm{v}_w|^2 C_D(\alpha (\hat{r})) \, \mathrm{sgn}(\bm{e}_c^\top\,\bm{v}_w)/2 \\
    \bm{f}_{a_L} &= \smallint_{0}^{1} (f_l(\hat{r}) \bm{e}_n - f_d(\hat{r})\bm{e}_c ) d\hat{r} \\
    \bm{\tau}_{a_L} &= \smallint_{0}^{1} \bm{l}_{f_L} (\hat{r}) \times (f_l(\hat{r}) \bm{e}_n - f_d(\hat{r})\bm{e}_c ) d\hat{r},
\end{aligned}
\label{eq:aerodynamic_formulation_2}   
\end{equation}
where $\rho$ is the air density, $\bm{v}_w$ is the airfoil velocity at $\bm{l}_{f_L}$, $\bm{e}_n$ and $\bm{e}_c$ are the unit vectors as shown in Fig. \ref{fig:wing_aerodynamics}. $C_L$ and $C_D$ are the lift and drag coefficients defined below:
\begin{equation}
\begin{aligned}
    C_L(\alpha) &= 0.225 + 1.58 \sin(2.13 \alpha - 7.2^\circ) \\
    C_D(\alpha) &= 1.92 - 1.55 \cos(2.04 \alpha - 9.82^\circ),
\end{aligned}
\label{eq:drag_lift_coefficients}   
\end{equation}
where $\alpha$ is the angle of attack in degrees. These coefficients are used in \cite{sane2001control} for a flapping wing MAV based on a fruit fly. The integration about the wing span (integrate about $\hat{c}$) and the right armwing side follow a similar derivation. Calculating the sum of all of this forces and torques forms the aerodynamic actuation vector $\bm{u}_a$.

\section{Gait Optimization Framework}
\label{sec:optimization}

\begin{table}[t]
    \vspace{0.05in}
    \centering
    \caption{List of the parameters and their values}
    \vspace{-0.1in}
    \begin{tabular}{c c | c c}
    \toprule
    Param. & Value & Param. & Value \\
    \midrule
    $m_B$ & 5 g & $\bm{l}_{L1}$ & $[0, 25, 25]^\top$ mm \\
    $m_A$ & 0.35 g & $\bm{l}_{L2}$ & $[0, 0, 50]^\top$ mm \\
    $m_W$ & 5.6 g & $\bm{l}_{L3}$ & $[0, 0, 150]^\top$ mm \\
    $\hat{I}_{B,x}$ & 0.625 g.cm$^2$ & $\bm{l}_{R1}$ & $[0, -25, 25]^\top$ mm \\
    $\hat{I}_{B,y}$ & 3.65 g.cm$^2$ & $\bm{l}_{R2}$ & $[0, 0, 50]^\top$ mm \\
    $\hat{I}_{B,z}$ & 3.65 g.cm$^2$ & $\bm{l}_{R3}$ & $[0, 0, 150]^\top$ mm \\
    $\hat{I}_{A,x}$ & 0.147 g.cm$^2$ & $\hat{I}_{W,x}$ & 1.05 g.cm$^2$ \\
    $\hat{I}_{A,y}$ & 0.147 g.cm$^2$ & $\hat{I}_{W,y}$ & 2.11 g.cm$^2$  \\
    $\hat{I}_{A,z}$ & 0.040 g.cm$^2$ & $\hat{I}_{W,z}$ & 2.11 g.cm$^2$  \\
    $w_c$ & 150 mm & $ \rho$ & 1 kg/m$^3$\\
    $w_r$ & 150 mm & $\Omega$ & 10 Hz \\ 
    \bottomrule
    \end{tabular}
    \label{tab:parameters_list}
    \vspace{-0.05in}
\end{table}

\begin{figure}[t]
    \vspace{0.05in}
    \centering
    \includegraphics[width=\linewidth]{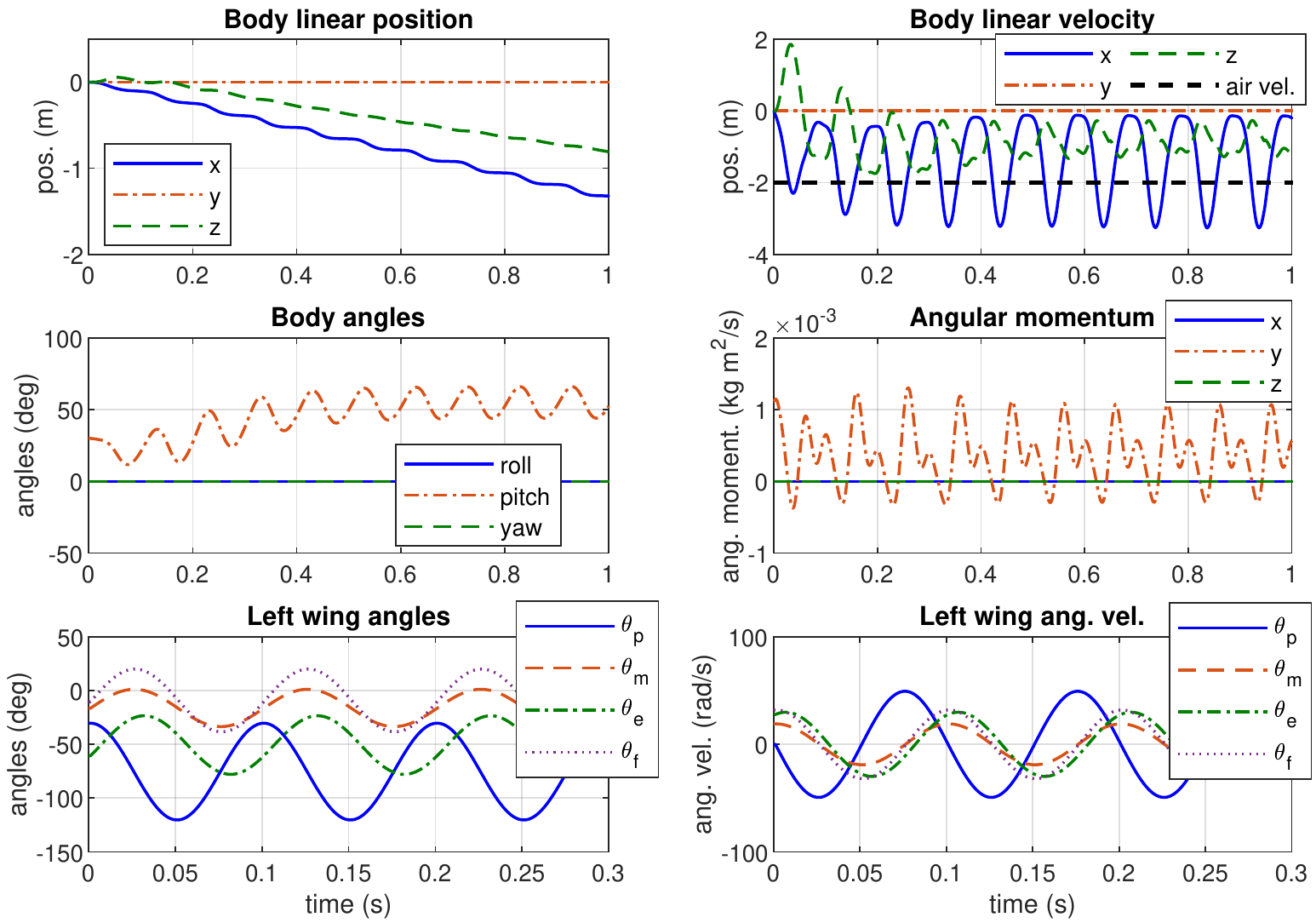}
    \caption{Open loop simulation for zero angular momentum gait using the optimized parameters in \eqref{eq:optimized_parameters}. The mean of the angular momentum is approximately zero and the linear velocity is approximately constant.}
    \label{fig:plot_gait_optimization}
    \vspace{-0.05in}
\end{figure}

This section outlines the optimization framework used in order to find the gait which stabilizes the robot's angular momentum and another gait for the upside-down perching maneuver. In order to define the angular momentum about the robot's center of mass, define the system center of mass:
\begin{equation}
    \bm{p}_{CoM} = \frac{1}{\sum_{F\in\mathcal{F}} m_F} \sum_{F\in \mathcal{F}} \bm{p}_F \, m_F.
\label{eq:center_of_mass}
\end{equation}
Then calculate the inertial angular momentum of the bodies about this center of mass:
\begin{equation}
\begin{aligned}
    \bm{\Pi}_{B} = & \, R_B\,\hat{I}_B\,\bm{\omega}_B^B - m_B\,(\bm{p}_B - \bm{p}_{CoM}) \times \dot{\bm{p}}_B \\
    \bm{\Pi}_{A_L} = & \, R_B\,R_{A_L}\,\hat{I}_{A_L}\,R_{A_L}^\top\, \bm{\omega}^{B}_{A_L} \\ & \qquad - m_A\,(\bm{p}_{A_L} - \bm{p}_{CoM}) \times \dot{\bm{p}}_{A_L} \\
    \bm{\Pi}_{W_L} = & \, R_B\,R_{A_L}\,R_{W_L}\,\hat{I}_{W_L}\,R_{W_L}^\top\, \bm{\omega}^{A_L}_{W_L}  \\ & \qquad - m_W\,(\bm{p}_{W_L} - \bm{p}_{CoM}) \times \dot{\bm{p}}_{W_L}.
\end{aligned}
\label{eq:angular_momentum_left}    
\end{equation}
Then the total angular momentum of the system about the center of mass is:
\begin{equation}
    \bm{\Pi} = \bm{\Pi}_{B} + \bm{\Pi}_{A_L} + \bm{\Pi}_{A_R} + \bm{\Pi}_{W_L} + \bm{\Pi}_{W_R}.
    \label{eq:angular_momentum}    
\end{equation}


In order to design the optimization problem to find the best gait for a stable flapping flight, we introduce a nonholonomic constraint at the wing joints acceleration so we can impose a specific gait for the armwing joints in the simulation. This can be implemented by using a Lagrangian multiplier to impose the acceleration constraint on \eqref{eq:eom_summary}, as shown below
\begin{equation}
\begin{aligned}
    M\,\dot{\bm{q}}_d + \bm{h} &= B_a\,\bm{u}_a + B_m\,\bm{u}_m + J_c^\top \, \bm{\lambda} \\
    J_c\,\dot{\bm{q}}_d &= \ddot{\bm{\theta}}_c,
\end{aligned}
\label{eq:accel_constraint}
\end{equation}
where $\ddot{\bm{\theta}}_c = [\ddot{\bm{\theta}}_{c,L}^\top, \ddot{\bm{\theta}}_{c,R}^\top]^\top$ is the joint acceleration constraint, $\bm{\theta}_{c,L}$ and $\bm{\theta}_{c,R}$ are the constraints for the left and right joint angles respectively, and $J_c = [0_{8 \times 6}, I_{8 \times 8}]$. Then the Lagrangian multiplier $\bm{\lambda}$ can be solved as follows:
\begin{equation}
    \bm{\lambda} = (J_c\,M^{-1}\,J_c^\top)^{-1}\, \left( J_c\,M^{-1}\,(\bm{h} - \bm{u}) - \ddot{\bm{\theta}}_c \right),
\label{eq:lagrangian_multiplier}
\end{equation}
where $\bm{u} = B_a \bm{u}_a + B_m \bm{u}_m$ is the combined forces acting on the system. The $\bm{u}_m$ value is irrelevant as the consequence of using the nonholonomic constraint on the joint angles.

\subsection{Zero Angular Momentum Gait Optimization}
\label{ssec:optimization_gait}

The following open loop trajectory is used for the the left armwing:
\begin{equation}
\begin{aligned}
    \bm{\theta}_{c,L} (t) = \begin{bmatrix}
        A_p \,\cos(\Omega \, t) + \thickbar{\theta}_{p}\\
        A_m \,\cos(\Omega \, t + \phi_m) + \thickbar{\theta}_{m}\\
        A_e \,\cos(\Omega \, t + \phi_e) + \thickbar{\theta}_{e}\\
        A_f \,\cos(\Omega \, t + \phi_f) + \thickbar{\theta}_{f}
    \end{bmatrix},
\end{aligned}
\label{eq:desired_trajectory}
\end{equation}
where $\Omega$ is the flapping frequency, $\thickbar{\theta}$ is the mean joint angle, $A$ is the amplitude, and $\phi$ is the phase shift of the other joints with respect to the plunging motion. The right armwing trajectory is symmetric to the left armwing which results in a symmetric aerodynamic force and stable in the roll and yaw motion. Then, there are 11 parameters to optimize using the following optimization problem:
\begin{equation}
\begin{aligned}
\min_{\bm{k}_1} &  \textstyle  \sum_i ( \mathrm{trace} (\bm{z}_i^\top\, Q \, \bm{z}_i)) 
    \\
    \text{subject to} & \,\, \bm k_{1, min} \leq \bm k_1 \leq \bm k_{1, max}, 
\end{aligned}
\label{eq:optimizer}
\end{equation}
\begin{equation}
\begin{aligned}
    \bm{k}_1 &= [\thickbar{\theta}_p, \thickbar{\theta}_m, \thickbar{\theta}_e, \thickbar{\theta}_f, A_p, A_m, A_e, A_f, \phi_m, \phi_e, \phi_f]^\top \\
    \bm{z} &= [\bm{\Pi}^\top,\,  \dot{p}_{B,z}]^\top,
\end{aligned}
\label{eq:optimizer_parameters}
\end{equation}
where $Q$ is a weighting matrix. The optimization cost function is the summation of the $\mathrm{trace} (\bm{z}_i^\top\, Q \, \bm{z}_i)$ in a numerical simulation of the wing dynamics where $i$ is the $i$'th time step of the numerical simulation. This cost function is designed to find the gait that can steadily keep the robot's angular momentum constant while also capable of maintaining zero vertical velocity throughout the simulation.

\subsection{Upside-down Perching Optimization}
\label{ssec:optimization_flip}

The trajectory found in \eqref{eq:optimizer} will be used to find the upside-down perching maneuver, which is done by introducing the appropriate force imbalances through the wing joints articulation. Since the gait found in \eqref{eq:optimizer} is momentum stable, we can simply superimpose a joint trajectory on top of this gait to create the force imbalance. We choose to superimpose a simple offset trajectory:
\begin{equation}
\begin{aligned}
\tfrac{d}{dt}\thickbar{\bm{\theta}}(t) = 
\begin{cases}
\bm{d}/T, & \text{if $t \geq t_0$ and $t < t_0 + T$} \\
-\bm{d}/T, & \text{if $t \geq t_0 + T$ and $t < t_0 + 2T$} \\
0, & \text{otherwise}
\end{cases},
\end{aligned}
\label{eq:offset_trajectory}
\end{equation}
where $\thickbar{\bm \theta} = [\thickbar{\theta}_p, \thickbar{\theta}_m, \thickbar{\theta}_e, \thickbar{\theta}_f]^\top$, $\bm{d} = [d_p, d_m, d_e, d_f]^\top$ is the optimization parameters which generates a triangular offset function with amplitude $\bm{d}$, ramp up and down period of $T$, and ramp startup time of $t_0$. To form a force imbalance, we set the left and right wing offset trajectory to be equal, i.e. $\tfrac{d}{dt}\thickbar{\bm{\theta}}_L(t) = \tfrac{d}{dt}\thickbar{\bm{\theta}}_R(t) = \tfrac{d}{dt}\thickbar{\bm{\theta}}(t)$. However, since the joint acceleration is the input to our constrained system, we need to impose a bounded acceleration for a smooth velocity trajectory. This is done by using a workaround where we close the loop by setting the acceleration constraint as
\begin{equation}
\ddot{\bm{\theta}}_c = -120 (\bm{\theta} - \bm{\theta}_r) - 120 (\dot{\bm{\theta}} - \dot{\bm{\theta}}_r),
\label{eq:optimization_2_constraints}
\end{equation}
where $\bm{\theta} = [\bm{\theta}_L^\top, \bm{\theta}_R^\top]^\top$, and $\bm{\theta}_r$ is the joint angle trajectories from \eqref{eq:desired_trajectory} superimposed with the offset trajectory from \eqref{eq:offset_trajectory}. The constraint in \eqref{eq:optimization_2_constraints} forms a simple PD controller which should give us a bounded joint accelerations. Finally, the objective function is to follow the trajectory of $\bm{\omega}^B_B$ that undertakes a rolling motion. The rolling motion is defined using the following function:
\begin{equation}
\begin{aligned}
\phi_r &= (\pi/2)\tanh(\eta(t)), \quad \eta(t) = (6/2T)(t - t_0) - 3 \\
\dot{\phi}_r &= (\pi/2)(3/2T)(1 - \tanh(\eta(t))^2),
\end{aligned}
\label{eq:target_angular_velocity}
\end{equation}
where $\phi_r$ is the target rolling angle. \eqref{eq:target_angular_velocity} forms a smooth ramp up of $180^\circ$ in roll angle within a time span of $2T$ starting from $t_0$. Then the optimization is setup as follows:
\begin{equation}
\begin{aligned}
    \min_{\bm{k}_2} & \, \textstyle \sum_i (\omega^B_{B,x} - \dot{\phi}_r)^2, 
    \qquad 
    \bm{k}_2 = [t_0, \bm{d}^\top]^\top\\
    \text{subject to} \, & \,\, \bm k_{2, min} \leq \bm k_2 \leq \bm k_{2, max}, 
\end{aligned}
\label{eq:optimizer_2}
\end{equation}
where $\omega^B_{B,x}$ is the first component of $\bm{\omega}_B^B$ which represents the body angular velocity about the roll axis, and $i$ is the simulation discrete time step between $t_0$ and $t_0 + 2T$.

\section{Simulation and Analysis}
\label{sec:simulation}

\begin{figure}[t]
    \vspace{0.05in}
    \centering
    \includegraphics[width=\linewidth]{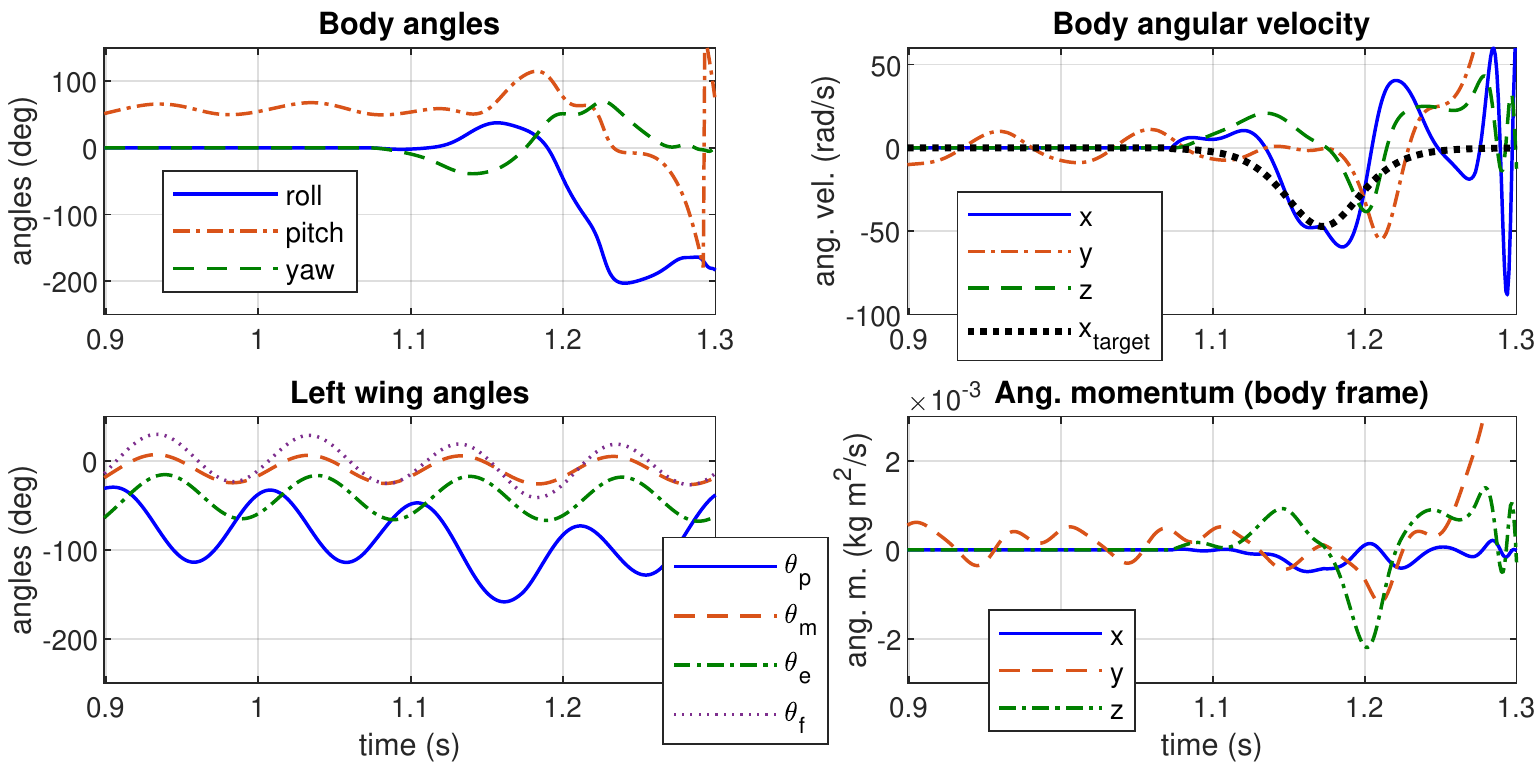}
    \caption{Open loop simulation for the upside-down perching using the optimized parameters from \eqref{eq:optimized_parameters} and \eqref{eq:optimized_parameters_2}. The robot follows the desired roll velocity but there is a significant overshoot during the slowing down period.}
    \label{fig:plot_perching_optimization}
    \vspace{-0.05in}
\end{figure}

\begin{figure}[t]
    \vspace{0.05in}
    \centering
    \includegraphics[width=\linewidth]{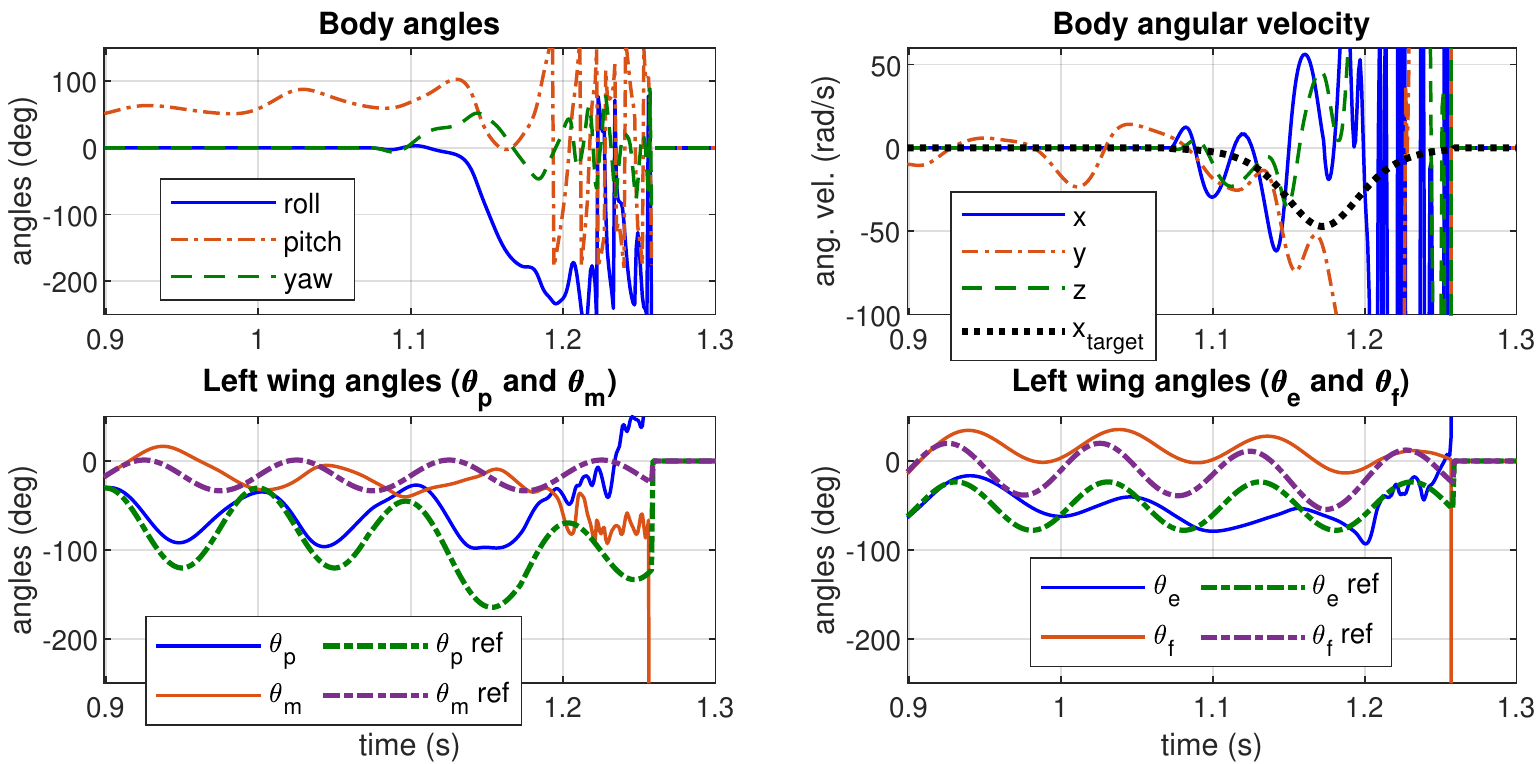}
    \caption{Closed loop simulation using PID controller to track the trajectories found in the optimizations. The roll angle reached the -180$^\circ$ for perching which indicates its feasibility. However, the controller failed to fully track the desired trajectory and the simulation is unstable after the upside-down maneuver.}
    \label{fig:plot_perching_control}
    \vspace{-0.05in}
\end{figure}

\begin{figure*}[t]
    \vspace{0.05in}
    \centering 
    \includegraphics[width=0.95\linewidth]{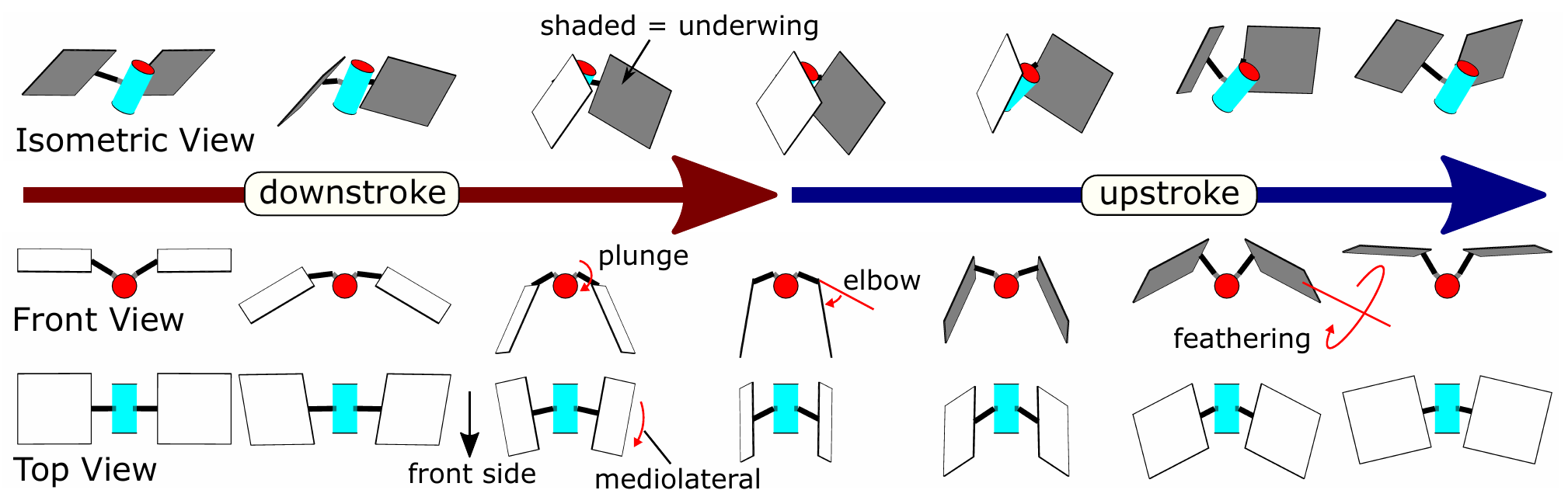}
    \caption{The Aerobat zero angular momentum flapping motion. The flapping motion is accompanied by the wing elbow expansion, mediolateral motion where the shoulder bends forwards, and the feathering motion which rotates the wing about the elbow joint. During the upstroke, the elbow joint is retracted and the feathering movement adjusts the angle of attack which reduces negative lift.}
    \vspace{-0.05in}
\label{fig:plot_flapping}
\end{figure*}

The optimization was run using RK4 algorithm where the robot simply ran in an open loop using the joint constraints described in \eqref{eq:accel_constraint} and \eqref{eq:desired_trajectory} for 2 seconds. The robot is subjected to a constant wind speed $\bm v_{air} = [-2,0,0]^\top$m/s about the inertial frame and the system parameters are listed in Table \ref{tab:parameters_list}. We use the weighting matrix of $Q = \mathrm{diag}([5, 5, 5, 10^{-5}])$ which gives a cost function ratio approximately equal between the momentum and the robot's vertical velocity. Using the interior-point algorithm, the optimization results in the following trajectory parameters:
\begin{equation}
\begin{aligned}
    \bm{k}_1 = [ & -75.3^\circ, -16.2^\circ, -50.7^\circ, -9.2^\circ, 45.0^\circ, 17.3^\circ,  \\
    & 27.2^\circ, 29.1^\circ, -91^\circ, -112^\circ, -92.5^\circ]^\top.
\end{aligned}
\label{eq:optimized_parameters}   
\end{equation}
The resulting gait and the robot's states can be seen in Fig. \ref{fig:plot_gait_optimization} and the illustration of this gait is shown in Fig. \ref{fig:plot_flapping}. The robot has a limit cycle with a stable angular and linear velocities with a mean pitch angle angle of $55^\circ$. However, if the force symmetry is lost due to the numerical error or other means, the robot starts to spin wildly and the open-loop gait is incapable of stabilizing the robot.

The optimization to find upside-down perching maneuver in \eqref{eq:optimizer_2} utilizes the gait in  \eqref{eq:optimized_parameters} to solve the optimization problem. We set $T = 0.2$s which is 2 wingbeats at 10 Hz flapping frequency. Then we bound $t_0 \in [1.0, 1.1]$s such that $t_0$ represents the start time of this maneuver within a wingbeat. The optimizer results in the following parameters:
\begin{equation}
    \bm{k}_2 = [1.0724 \text{ s}, -55.7^\circ, 0^\circ, 0^\circ, -16.6^\circ]^\top.
    \label{eq:optimized_parameters_2}
\end{equation}
The $t_0$ of 1.0724 seconds indicates that the maneuver starts at 72.4\% of a wingbeat period. Interestingly, the mediolateral and elbow offsets are both zero even though we did not constrain the optimizer this way. This indicates that the plunging and feathering movements are the primary driving force for the upside-down maneuver. Fig. \ref{fig:aerobat_flip} illustrates the gait found in this optimization and Fig. \ref{fig:plot_perching_optimization} shows the simulation states of the robot using the optimized parameters where it shows that the body's Euler angles does a $180^\circ$ turn in approximately 0.15 seconds. The body angular velocity follows the target velocity for a brief period of time but then overshoots during the ramping down period. The robot spins out of control after 1.3 seconds mark due to its upside-down state, but we can assume that it already has perched onto something by then using our launching landing gear \textit{Harpoon} \cite{ramezani2020towards}.

Now that we have identified all the trajectories to follow, we can then use a simple PID controller to track these target trajectory and see if the result matches what the optimizer predicted. We use PID controller gains of $k_p = k_d = 0.0012$ and $k_i = 0.006$. Fig. \ref{fig:plot_perching_control} shows the controller performance, where the robot achieved the upside-down perching maneuver but very quickly turned unstable. Fig. \ref{fig:plot_perching_control} also shows that the controller does not track the desired trajectory well, which is likely caused by the PID controller's incapability to track the desired trajectory in the presence of the changes in aerodynamic forces and gravity. This result indicates that a constant PID gain is not sufficient to track the trajectory and we need to develop a better controller. Additionally, a momentum stabilizing controller also need to be implemented so that the robot can perform the upside-down maneuver more stably.

\section{Conclusions and Future Work}
\label{sec:conclusion}

\balance

A dynamic model of a bat robot and the optimization framework using nonholonomic constraint to find the gait which is momentum stable followed by an upside-down perching maneuver is presented in this paper. The optimizer has successfully found the parameters which fulfills our criterion and the corresponding gaits for the stable flying and upside-down perching has been found. However, the PID controller we used to track these trajectory is insufficient and we need to improve the controller's performance to track these trajectory. For the future work, we can investigate the implementation of a robust trajectory tracking controller using a rapidly exponentially stable controllers, such as \cite{ames2014rapidly}, and a nonlinear controller which actively stabilizes the robot's angular momentum. Additionally, we can design a \textit{kinetic sculpture} that incorporate these DoFs and investigate a \textit{morphological computation} framework that might be more natural for a bio-inspired flexible armwing mechanism.





\bibliographystyle{IEEEtran}
\bibliography{references.bib}

\end{document}